\documentclass[conference]{IEEEtran}

\usepackage{cite}
\usepackage{amsmath,amssymb,amsfonts}
\usepackage{algorithmic}
\usepackage{algorithm}
\usepackage{multirow}
\usepackage{graphicx}
\usepackage{svg}
\usepackage{textcomp}
\usepackage{xcolor, soul}
\usepackage{booktabs}
\usepackage{mathrsfs}

\def\BibTeX{{\rm B\kern-.05em{\sc i\kern-.025em b}\kern-.08em
    T\kern-.1667em\lower.7ex\hbox{E}\kern-.125emX}}
\begin{document}

\title{VS-Graph: Scalable and Efficient Graph Classification Using Hyperdimensional Computing}

\author{
\IEEEauthorblockN{
Hamed Poursiami\IEEEauthorrefmark{1}, 
Shay Snyder\IEEEauthorrefmark{1}, 
Guojing Cong\IEEEauthorrefmark{2}, 
Thomas Potok\IEEEauthorrefmark{2}, 
Maryam Parsa\IEEEauthorrefmark{1}\IEEEauthorrefmark{3}}
\IEEEauthorblockA{\IEEEauthorrefmark{1}Department of Electrical and Computer Engineering, George Mason University, Fairfax, VA, USA}
\IEEEauthorblockA{\IEEEauthorrefmark{2}Oak Ridge National Laboratory, Oak Ridge, TN, USA}
\IEEEauthorblockA{\IEEEauthorrefmark{3}Corresponding Author}
}

\maketitle
\begin{abstract}

Graph classification is a fundamental task in domains ranging from molecular property prediction to materials design. While graph neural networks (GNNs) achieve strong performance by learning expressive representations via message passing, they incur high computational costs, limiting their scalability and deployment on resource-constrained devices. Hyperdimensional Computing (HDC), also known as Vector Symbolic Architectures (VSA), offers a lightweight, brain-inspired alternative, yet existing HDC-based graph methods typically struggle to match the predictive performance of GNNs. In this work, we propose VS-Graph, a vector-symbolic graph learning framework that narrows the gap between the efficiency of HDC and the expressive power of message passing. VS-Graph introduces a Spike Diffusion mechanism for topology-driven node identification and an Associative Message Passing scheme for multi-hop neighborhood aggregation entirely within the high-dimensional vector space. Without gradient-based optimization or backpropagation, our method achieves competitive accuracy with modern GNNs, outperforming the prior HDC baseline by 4-5\% on standard benchmarks such as MUTAG and DD. It also matches or exceeds the performance of the GNN baselines on several datasets while accelerating the training by a factor of up to $450\times$. Furthermore, VS-Graph maintains high accuracy even with the hypervector dimensionality reduced to $D=128$, demonstrating robustness under aggressive dimension compression and paving the way for ultra-efficient execution on edge and neuromorphic hardware.
\end{abstract}


\section{Introduction}
\label{sec:intro}

Graph-structured data arise in a wide range of scientific and industrial domains, including molecular chemistry~\cite{zhao2024molecular}, social and information networks~\cite{sharma2024survey,liu2024sbp}, recommendation systems~\cite{wu2022graph}, and materials science~\cite{merchant2023scaling}. In these settings, entities and their relationships are naturally modeled as nodes and edges in a graph that resides in a non-Euclidean domain, where the notion of locality is determined by connectivity rather than spatial proximity. Graph learning is commonly organized into three canonical task families: node-level tasks, such as semi-supervised (transductive) classification in citation networks, which infer labels for individual nodes given only small labeled subsets~\cite{wu2022nodeformer}; edge-level tasks, including link prediction in recommender systems~\cite{yu2022graph}, which estimate the presence or type of relations between pairs of nodes; and graph-level tasks, which map an entire graph (e.g., a molecule or crystalline material) to a label that reflects its functional or physical properties~\cite{liu2022spherical}. Across these settings, a central challenge is to develop representations that capture expressive structural information while enabling scalable and computationally efficient learning~\cite{kose2024survey}.

Graph Neural Networks (GNNs) have become a widely used framework for such problems~\cite{gkarmpounis2024survey}. They are typically formulated as message passing architectures in which nodes iteratively aggregate and transform information from their neighbors through a sequence of layers. Building on the Message Passing Neural Network (MPNN)~\cite{gilmer2017neural} formulation, subsequent models such as Graph Convolutional Networks (GCN)~\cite{kipf2017semi}, Graph Attention Networks (GAT)~\cite{velickovic2018graph}, Graph Isomorphism Networks (GIN)~\cite{xu2019how}, and Principal Neighborhood Aggregation (PNA)~\cite{corso2020principal}, instantiate this paradigm through different choices of aggregation functions, normalization schemes, and update mechanisms. While these models achieve strong empirical performance, they face practical challenges related to scalability and efficiency~\cite{wu2025survey}. Training  these GNNs requires backpropagation over sparse and irregular graph structures, which leads to high memory consumption and inefficient use of parallel hardware~\cite{hamilton2017inductive,wu2025survey}. In addition, deeper architectures are prone to over-smoothing and incur higher computational cost that limits deployment on resource-constrained devices~\cite{li2018deeper,li2025scalegnn}.

Hyperdimensional Computing (HDC), also known as Vector Symbolic Architectures (VSA)~\cite{plate1995holographic}, is a brain-inspired computational framework that encodes information using high-dimensional distributed hypervectors~\cite{kanerva2009hyperdimensional}. HDC operates via a set of algebraic operations, including binding, bundling, and permutation, which enable the construction of complex symbolic structures from atomic representations and support robust, noise-tolerant, and highly parallel computation~\cite{kleyko2022survey}. The efficiency and representational flexibility of HDC have motivated recent efforts to extend it to graph learning.

At the node level, methods such as HDGL~\cite{dalvi2025hyperdimensional} and HyperGraphX ~\cite{cong2025hypergraphx} show that lightweight hyperdimensional operations can generate expressive node embeddings from local structural information, enabling efficient inference with minimal training overhead. Related work such as RelHD~\cite{kang2022relhd} further highlights the compatibility of HDC with energy-efficient processing-in-memory hardware, illustrating the potential for low-power acceleration of graph learning workloads. Parallel efforts have investigated HDC for graph-level tasks, including molecular property prediction~\cite{ma2022molehd} and materials classification~\cite{gobin2024exploration}, where hyperdimensional representations are constructed to capture whole-graph structure for downstream learning. Within this line of work, GraphHD~\cite{nunes2022graphhd} provides a general framework for graph classification by forming edge-level hypervectors through binding of structurally informed node representations and aggregating them into a single holistic embedding. These studies demonstrate the promise of HDC as an efficient and scalable direction for graph classification.

Despite these advancements, existing HDC-based methods still exhibit a performance gap compared to state-of-the-art GNNs on standard graph classification benchmarks~\cite{gobin2024exploration,verges2024molecular,nunes2022graphhd}. While HDC approaches typically offer notable computational efficiency, they often struggle to match the predictive fidelity of models with learnable message passing layers~\cite{dalvi2025hyperdimensional}. In this work, we propose \textbf{VS-Graph}, a novel vector-symbolic graph learning framework that integrates the computational efficiency of high-dimensional algebra with the structural expressiveness of message passing. Unlike traditional GNNs that require training large numbers of parameters through iterative backpropagation, our approach employs a weight-free multi-hop aggregation mechanism that operates entirely within the hyperdimensional space, allowing the model to capture long-range structural context in a single encoding pass. Our specific contributions are as follows:

\begin{enumerate}

    \item \textbf{Vector-symbolic graph encoder.} We develop a vector-symbolic graph classification framework that combines \textit{``Spike Diffusion"}, a novel topology-derived node identification mechanism, with \textit{``Associative Message Passing"}, a lightweight multi-hop aggregation scheme that composes structural information entirely within hyperdimensional space.

    \item \textbf{Improved performance–efficiency trade-off.} We show that our framework substantially improves over the HDC baseline by margins of up to $5\%$ and achieves accuracy competitive with standard GNN baselines, while offering ultra-low inference latency and significantly reduced training cost by a factor of up to $450\times$. 

    \item \textbf{Robustness under dimensionality reduction.} We demonstrate that our method maintains strong predictive performance even with hypervector dimensionality reduced from $D = 8192$ to $D = 128$, with less than $1.5\%$ accuracy degradation, facilitating efficient deployment on resource-constrained edge platforms and emerging neuromorphic hardware.

\end{enumerate}

\section{Preliminaries and Related Work}
\label{sec:preliminary}

A graph classification dataset is defined as a collection $\mathcal{D} = \{(G_1, y_1), \dots, (G_N, y_N)\}$. Each instance consists of a graph $G_i = (V_i, E_i)$, comprised of a node set $V_i$ and an edge set $E_i$, paired with a corresponding class label $y_i$ from a label space $\mathcal{Y}$. The objective of graph classification is to learn a mapping function $f: \mathcal{G} \to \mathcal{Y}$ that accurately predicts the label for an unseen graph. 

\subsection{Graph Neural Networks (GNNs)}

GNNs have emerged as a widely adopted framework for addressing the graph classification task. These methods typically follow the message passing neural network paradigm~\cite{gilmer2017neural}, where node representations are iteratively updated by aggregating features from local neighborhoods. Formally, the representation of node $v$ at layer $l$ is updated as:

\begin{equation} h_v^{(l)} = \Phi \left( h_v^{(l-1)}, \text{AGG}\left({h_u^{(l-1)} : u \in \mathcal{N}(v)}\right) \right) \end{equation}

where $\mathcal{N}(v)$ denotes the set of neighbors of $v$. Here, $\text{AGG}$ represents a permutation-invariant aggregation function that compiles neighborhood information, and $\Phi$ denotes a differentiable update function that combines the aggregated message with the node's previous state. We consider three prominent GNN implementations that utilize this framework:

\paragraph{Graph Convolutional Networks (GCN)} GCN~\cite{kipf2017semi} performs a localized first-order approximation of spectral graph convolutions. It applies a fixed symmetric normalization to the adjacency matrix during aggregation, computing a weighted average of neighbor features to update node embeddings.

\paragraph{Graph Attention Networks (GAT)} GAT~\cite{velickovic2018graph} introduces an attention mechanism to the aggregation process. This architecture allows the model to learn and assign different importance weights to neighbors within the graph structure, enabling the aggregation function to prioritize specific nodes based on their features.

\paragraph{Graph Isomorphism Networks (GIN)} GIN~\cite{xu2019how} employs an aggregation scheme designed to maximize discriminative power, ensuring the model is as expressive as the Weisfeiler–Lehman (WL) graph isomorphism test~\cite{shervashidze2011weisfeiler}. It uses a sum aggregator together with a Multi-Layer Perceptron (MLP) to learn injective mappings over the multiset of neighbor features, and incorporates a learnable parameter $\epsilon$ that controls the balance between a node’s own features and those of its neighbors.

\subsection{Hyperdimensional Computing (HDC)}
HDC, also known as Vector Symbolic Architectures (VSA), is a brain-inspired computational framework that represents information using very high-dimensional vectors ($D \ge 10,000$) called hypervectors. In such spaces, randomly generated hypervectors are nearly orthogonal (pseudo-orthogonal), enabling robust, distributed representation of information~\cite{plate1995holographic}. HDC operates in binary $\{0,1\}^D$ or bipolar $\{-1,1\}^D$ spaces and relies on a small set of algebraic operations:

\begin{itemize}
    \item \textbf{Bundling ($\oplus$):} Combines multiple hypervectors into a single representative vector that remains similar to its inputs. It is typically implemented via element-wise majority voting (binary) or summation with thresholding (bipolar).
    \item \textbf{Binding ($\otimes$):} Associates two hypervectors to form a new vector dissimilar from both inputs, typically implemented via XOR (binary) or element-wise multiplication (bipolar).
    \item \textbf{Similarity ($\delta$):} Matching and inference are performed by measuring distances between hypervectors, commonly using Hamming distance  or cosine-similarity.
\end{itemize}

Together, these operations provide a robust and flexible algebra over high-dimensional vectors, supporting the construction, combination, and comparison of distributed representations~\cite{verges2025classification}.

\paragraph{GraphHD}

GraphHD~\cite{nunes2022graphhd} is a hyperdimensional approach designed for graph classification that maps topological structure directly to hypervectors. It uses PageRank~\cite{page1999pagerank} to derive a structural signature for each node and discretizes the PageRank scores into rank bins. Each rank bin is assigned a unique random hypervector from a pre-generated basis memory. Once nodes are encoded, edges are represented by binding the hypervectors of their constituent nodes. Finally, the entire graph is represented by bundling all edge hypervectors together. This produces a single hypervector summarizing the graph’s structure. During training, graph hypervectors belonging to the same class are bundled into class prototypes, and inference assigns a new graph to the class whose prototype is most similar to its encoded representation.
\section{Methodology}
\label{sec: methodology}

Given a graph $G = (V, E)$, our primary objective is to learn a mapping function $f: G \rightarrow \mathbb{R}^D$ that projects the graph into a fixed-dimensional embedding space suitable for classification. Our framework operates strictly on the graph connectivity structure, intentionally disregarding any auxiliary information such as node or edge attributes. This design reflects the fact that such information may be unavailable, incomplete, or not directly comparable across different graph domains~\cite{hu2020open}. In contrast, the structural relations encoded by the adjacency pattern are always well defined within each graph. Focusing solely on topology also provides a controlled setting for assessing the representational capabilities of hyperdimensional methods on graphs, following the same structural principle used in GraphHD \cite{nunes2022graphhd}. We achieve this mapping through several successive components, including spike diffusion, associative message passing, graph level readout, and prototype classification, detailed below.

\begin{algorithm}[ht]
\caption{Vector Symbolic Graph Classification Pipeline}
\label{alg:method}
\begin{algorithmic}[1]
\REQUIRE Training set $\mathcal{D_G} = \{(G_k, y_k)\}$, global rank basis $B[r] \in \{0,1\}^D$, diffusion hops $K$, message passing layers $L$, blend factor $\alpha$

\COMMENT{\textbf{Phase 1: Graph Embeddings}}
\FOR{each graph $G_k = (V_k, E_k)$ in $\mathcal{D}$}
    \STATE Initialize unit spikes $s_i \leftarrow 1$ for all $i \in V_k$ 
    
    \FOR{$t = 1$ \TO $K$} 

        \STATE \COMMENT{Spike Diffusion}
    
        \FOR{each node $i \in V_k$}
    
            \STATE $s_i \leftarrow \sum_{j \in \mathcal{N}(i)} s_j$
        \ENDFOR
    \ENDFOR

    \STATE Obtain ranks $r_i \leftarrow \textsc{Rank}(s_i; \{s_j\}_{j \in V_k})$ for all $i \in V_k$
    \FOR{each node $i \in V_k$}
        \STATE $h_i^{(0)} \leftarrow B[r_i]$ \COMMENT{Binary hypervector for rank $r_i$}
    \ENDFOR

    \FOR{$\ell = 0$ \TO $L-1$}
    \STATE \COMMENT{Associative Message Passing}
        \FOR{each node $i \in V_k$}
            \STATE $m_i^{(\ell)} \leftarrow \bigvee_{j \in \mathcal{N}(i)} h_j^{(\ell)}$ \COMMENT{Logical OR aggregation}
            \STATE $h_i^{(\ell+1)} \leftarrow \alpha h_i^{(\ell)} + (1 - \alpha) m_i^{(\ell)}$  \COMMENT{Residual blend}
        \ENDFOR
    \ENDFOR

    \STATE $z_{G_k} \leftarrow \frac{1}{|V_k|} \sum_{i \in V_k} h_i^{(L)}$ \COMMENT{Graph level readout}
\ENDFOR

\COMMENT {\textbf{Phase 2: Prototype Construction}}
\FOR{each class $c$}
    \STATE Let $\mathcal{D}_c = \{G_k \mid y_k = c\}$ and $N_c = |\mathcal{D}_c|$
    \STATE $p_c' \leftarrow \frac{1}{N_c} \sum_{G_k \in \mathcal{D}_c} z_{G_k}$
    \STATE $p_c \leftarrow p_c' / \left(\|p_c'\|_2 + \varepsilon\right)$
\ENDFOR

\COMMENT{\textbf{Phase 3: Inference for a New Graph $G$}}
\STATE Compute $z_G$ from $G$ using Phase 1 
\STATE $z_G \leftarrow z_G / \left(\|z_G\|_2 + \varepsilon\right)$
\STATE $\hat{y} \leftarrow \arg\max_{c} \; z_G^{\top} p_c$

\end{algorithmic}
\end{algorithm}

\subsection{Spike Diffusion}
A fundamental challenge in graph representation learning is the absence of a trivial correspondence between nodes in different graphs. In images, for example, the pixel at position $(i,j)$ always refers to the same spatial location in the grid, while nodes in different graphs are completely anonymous~\cite{ling2023graph}; there is no inherent notion that “node $v$ in one graph” corresponds to “node $u$ in another.” To associate nodes with consistent identifiers that reflect their structural roles, we must first derive a signature from the graph’s topology.

We introduce a lightweight process, \textit{\textbf{``Spike Diffusion"}}, to compute such topology-derived signatures. Each node begins with a unit activation spike and propagates this spike outward through the graph. At every diffusion hop, a node integrates the spikes received from its neighbors and forwards the resulting signal to them in the next hop. Repeating this process for a fixed number of hops yields a diffusion response for each node, capturing how the node participates in the multi-hop flow of spikes through the graph.

These diffusion responses provide a natural way to impose an ordering on the nodes. Sorting nodes according to their responses establishes an ordinal ranking, which serves as a simple yet effective topology-based identifier. Finally, we map these rank-derived identifiers into hyperdimensional space by assigning a unique random binary hypervector to each rank index. In effect, nodes across different graphs that occupy comparable structural positions (i.e., that receive the same rank) are represented by the same binary hypervector.

\subsection{Associative Message Passing}

After obtaining rank based binary hypervectors from the \textit{``Spike Diffusion"} stage, we further refine the node level representations through \textit{\textbf{``Associative Message Passing"}}. Let $h_{i}^{(0)} \in \{0,1\}^{D}$ denote the hypervector assigned to node 
$i$ after the ranking and hypervector mapping. We apply $L$ layers of message passing to incorporate structural context.
For each layer $l = 0,..., L-1$ node $i$ aggregates information from its neighbors $\mathscr{N}(i)$ using a dimension-wise logical OR operator:
\begin{equation}
    m_{i}^{(l)} = \bigvee_{j \in \mathscr{N}(i)} h_{j}^{(l)} 
\end{equation}
A key property of this operator is its idempotency, meaning that repeated contributions from the same neighbors do not accumulate. This keeps the aggregation inherently stable and bounded, and removes the need for additional normalization.

After aggregating the neighborhood information, the update incorporates a residual contribution from the previous node state. The new representation is computed as a convex combination of the earlier state and the aggregated message:
\begin{equation}
    h_{i}^{(l+1)} = \alpha h_{i}^{(l)} + (1-\alpha)m_{i}^{(l)} 
\end{equation}

where $\alpha \in [0,1]$ determines the relative influence of the previous representation. After $L$ layers, the set $\{ h_{i}^{(L)} \}$ constitutes the final node-level encoding obtained through the message passing process.

\subsection{Graph-Level Readout and Prototype Classification}

To produce a single, graph-level embedding $z_G$ from the final node representations $\{ h_{i}^{(L)} \}$, we apply a global readout operation, implemented as an element-wise mean over all node hypervectors in the graph:
\begin{equation}
    z_G = \frac{1}{|V|} \sum_{i \in V} h_i^{(L)}
\end{equation} 
which yields a fixed $D$-dimensional embedding $z_G \in \mathbb{R}^D$ that serves as the final representation of the entire graph.

This graph-level embedding is then used for \textbf{prototype classification}, a non-parametric method consisting of two phases:
\paragraph{Prototype Construction (Training)} For each graph in the training set, we compute its embedding $z_i$. A single prototype vector $p_c$ is formed for each class $c$ by averaging the embeddings of all training graphs belonging to that class:
\begin{equation}
    p_c' = \frac{1}{N_c} \sum_{G \in D_c} z_G
\end{equation}
and normalizing the result:
\begin{equation}
    p_c = \frac{p_c'}{\left\| p_c' \right\|_2 + \epsilon} 
\end{equation}

\paragraph{Prototype Matching (Inference)} Given a graph at inference time, we compute its embedding $z_{test}$ and normalize it. The predicted label is obtained by selecting the prototype with maximum cosine similarity:
\begin{equation}
    \hat{y} = \arg \max_{c}   \left( z_{test} \cdot p_c \right)
\end{equation}

The complete computational workflow, integrating spike diffusion, associative message passing, graph level readout, and prototype-based classification, is outlined in Algorithm~\ref{alg:method}.

\section{Experimental Setup and Methods}
\label{sec:experiments}

\begin{figure*}[htbp]
  \centering
  \includegraphics[width=0.95\linewidth]{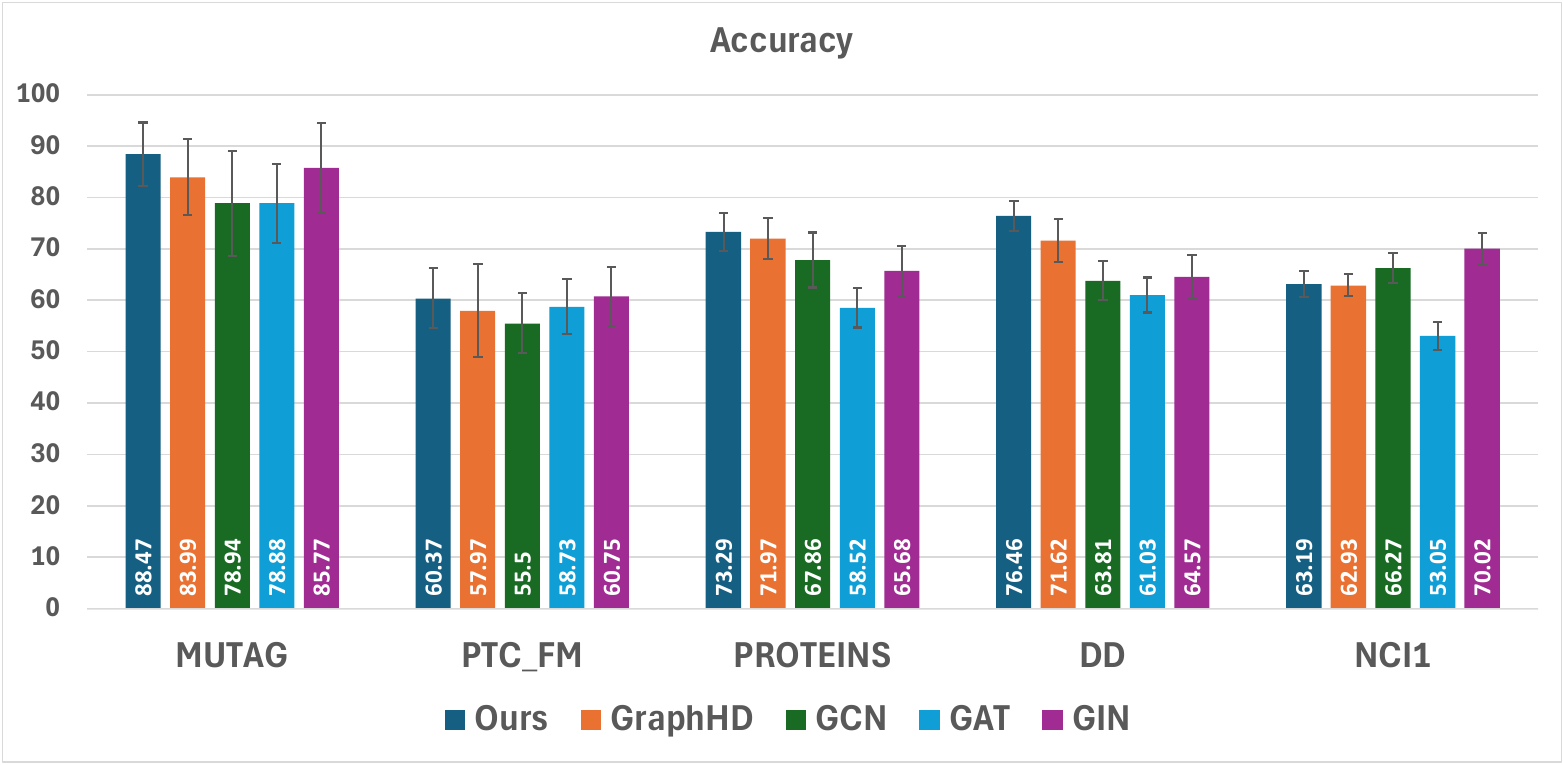}
  \caption{\textbf{Classification Accuracy Comparison Across Benchmark Datasets.} Predictive performance of VS-Graph (Ours) compared to the HDC baseline (GraphHD) and standard GNN models (GCN, GAT, GIN) across five benchmark datasets: MUTAG, PTC\_FM, PROTEINS, DD, and NCI1.}
  \label{fig:acc_overall}
\end{figure*}

We evaluate the proposed framework on several benchmark graph classification datasets and compare its performance with both hyperdimensional computing and graph neural network baselines. All experiments are implemented in PyTorch~\cite{paszke2019pytorch} with DGL~\cite{wang2019deep} as the graph processing library. Computations are performed using an NVIDIA Tesla T4 GPU, an Intel Xeon 2.20 GHz CPU, and 52 GB RAM.

\subsection{Datasets}

Our experiments use five widely adopted graph classification datasets from the TUDataset collection~\cite{morris2020tudataset}. These include MUTAG~\cite{debnath1991structure}, PTC\_FM~\cite{helma2001predictive}, PROTEINS~\cite{dobson2003distinguishing}, DD~\cite{shervashidze2011weisfeiler}, and NCI1~\cite{wale2008comparison}, each providing graphs paired with categorical graph labels. MUTAG and PTC\_FM contain small chemical compound graphs, PROTEINS consists of protein graphs where nodes represent amino acids, DD includes protein structures with larger and denser graphs, and NCI1 contains chemical compounds screened for anticancer activity. All datasets are treated in their topology-only form, relying solely on the adjacency structure. A summary of their key statistics is provided in Table~\ref{tab:datasets}.

\begin{table}[h]
\centering
\caption{Statistics of graph datasets}
\label{tab:datasets}
\begin{tabular}{lcccc}
\toprule
\textbf{Dataset} & \textbf{Graphs} & \textbf{Classes} & \textbf{Avg. vertices} & \textbf{Avg. edges} \\
\midrule
MUTAG     & 188  & 2 & 17.93  & 19.79  \\
PTC\_FM   & 349  & 2 & 14.11  & 14.48  \\
PROTEINS  & 1113 & 2 & 39.06  & 72.82  \\
DD        & 1178 & 2 & 284.32 & 715.66 \\
NCI1      & 4110 & 2 & 29.87  & 32.30 \\
\bottomrule
\end{tabular}
\end{table}

\subsection{Baselines and Evaluation Protocol}

For the hyperdimensional computing baseline, we include GraphHD~\cite{nunes2022graphhd} evaluated under the same dimensional settings as our method. For the GNN baselines, we consider GCN~\cite{kipf2017semi}, GAT~\cite{velickovic2018graph}, and GIN~\cite{xu2019how}, each configured to use only the connectivity structure without node or edge attributes to match our topology-only setting.

Evaluation follows 10-fold cross-validation, repeated three times with stratified splits. We report classification accuracy, average training time per graph, and average inference latency per graph, measured as wall clock time. For the hyperdimensional methods, we evaluate vector sizes from $2^{7}$ to $2^{13}$. Comparisons against GNN baselines are reported using the dimensionality $D=2^{13}=8192$.




\section{Results and Discussion}
\label{sec:discussion}

\begin{table}[ht]
\centering
\caption{Training time per graph (milliseconds)}
\label{tab:train_overall}
\resizebox{\columnwidth}{!}{%
\begin{tabular}{@{}lccccc@{}}
\toprule
 &
  \multicolumn{1}{l}{\textbf{Ours}} &
  \multicolumn{1}{l}{\textbf{GraphHD}} &
  \multicolumn{1}{l}{\textbf{GCN}} &
  \multicolumn{1}{l}{\textbf{GAT}} &
  \multicolumn{1}{l}{\textbf{GIN}} \\ \midrule
\textbf{MUTAG}    & 0.142 & 0.801 & 61.21 & 41.77 & 47.20 \\
\textbf{PTC\_FM}  & 0.153 & 0.929 & 36.71 & 33.93 & 37.39 \\
\textbf{PROTEINS} & 0.230 & 0.904 & 37.76 & 27.47 & 49.24 \\
\textbf{DD}       & 2.120 & 1.892 & 41.36 & 28.85 & 72.92 \\
\textbf{NCI1}     & 0.192 & 0.768 & 56.40 & 20.40 & 85.84 \\ \bottomrule
\end{tabular}%
}

\end{table}



The experimental results demonstrate that the proposed framework successfully reconciles the trade-off between computational efficiency and representational power in graph classification. In terms of predictive performance, as shown in Figure~\ref{fig:acc_overall}, our method consistently outperforms the hyperdimensional baseline, GraphHD, across all evaluated datasets, most notably in MUTAG and DD where we observe a 4\% to 5\% accuracy advantage. Compared to GNN baselines, our method achieves superior accuracy on MUTAG, PROTEINS, and DD, while delivering comparable performance on PTC\_FM. The only scenario where a baseline notably exceeds our performance is with GIN on the NCI1 dataset.

However, this accuracy comes at a substantial computational cost during training. As summarized in Table~\ref{tab:train_overall}, GIN requires approximately 85.8 milliseconds to train per graph on NCI1, whereas our method requires only 0.19 milliseconds, representing a speedup factor of nearly 450$\times$. For the full NCI1 dataset comprising over 4,000 graphs, this translates to a training time of more than five minutes for GIN compared to less than a second for our approach. This efficiency stems from the weight-free nature of our algorithm which eliminates iterative gradient-based optimization in favor of a single pass over the data. As shown in Table~\ref{tab:infer_overall}, this efficiency extends to inference where our method achieves comparable or superior latency across most datasets, maintaining real-time capability.

\begin{table}[ht]
\centering
\caption{Inference time per graph (milliseconds)}
\label{tab:infer_overall}
\resizebox{\columnwidth}{!}{%
\begin{tabular}{@{}lccccc@{}}
\toprule
 &
  \multicolumn{1}{l}{\textbf{Ours}} &
  \multicolumn{1}{l}{\textbf{GraphHD}} &
  \multicolumn{1}{l}{\textbf{GCN}} &
  \multicolumn{1}{l}{\textbf{GAT}} &
  \multicolumn{1}{l}{\textbf{GIN}} \\ \midrule
\textbf{MUTAG}    & 0.366 & 0.980  & 0.948 & 0.727 & 0.641 \\
\textbf{PTC\_FM}  & 0.368 & 1.143 & 0.572 & 0.492 & 0.552 \\
\textbf{PROTEINS} & 0.418 & 1.050 & 0.281 & 0.368 & 0.405 \\
\textbf{DD}       & 2.288 & 2.028 & 0.332 & 0.431 & 0.521 \\
\textbf{NCI1}     & 0.328 & 0.923 & 0.286 & 0.394 & 0.374 \\ \bottomrule
\end{tabular}%
}
\end{table}

\begin{figure*}[htbp]
  \centering
  \includegraphics[width=\linewidth]{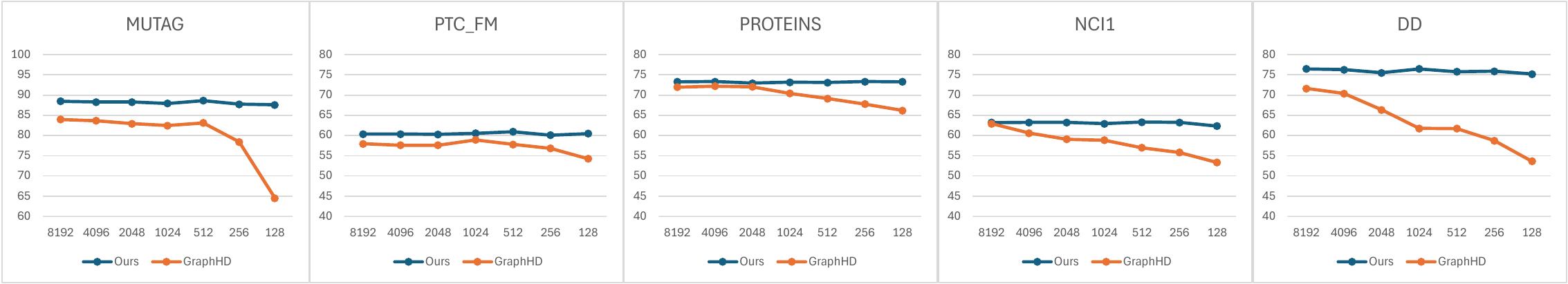}
  \caption{Effect of hypervector dimensionality on classification accuracy for VS-Graph (Ours) and GraphHD. Each subplot reports test accuracy as the hypervector size D is reduced from 8192 to 128, showing that VS-Graph remains robust under aggressive dimensionality reduction while GraphHD degrades more sharply.}
  \label{fig:acc_size}
\end{figure*}

An exception to this latency advantage is observed on the DD dataset which contains significantly larger graphs. Here, the overhead of associative message passing leads to increased computation time relative to the baselines. However, we can mitigate this by leveraging the model's robustness to dimensionality reduction. Figure~\ref{fig:acc_size} illustrates the classification accuracy across varying hypervector dimensions ($D$). We observe that our method maintains robust predictive performance even as dimensions are reduced to $D=128$, whereas GraphHD degrades significantly. This suggests that our encoding scheme is able to construct distinct, well-separated class prototypes even within lower-dimensional hyperspaces.

\begin{figure*}[htbp]
  \centering
  \includegraphics[width=\linewidth]{Figs/Latency_Size.pdf}
  \caption{\textbf{Training time} and \textbf{Inference latency} per graph (in milliseconds) for VS-Graph (Ours) and GraphHD versus hypervector dimensionality.}
  \label{fig:latency_size}
\end{figure*}

Figure~\ref{fig:latency_size} presents the training and inference latency trends with respect to vector size. While the resulting speedup is negligible for small graphs which are already processed rapidly, it becomes impactful for the larger graphs in DD. By reducing vector dimensions, we regain superior computational efficiency on complex graphs while retaining predictive performance. This flexibility is particularly advantageous for specialized hardware implementations where memory bandwidth and on-chip storage are limiting factors~\cite{trensch2022system}, enabling efficient neuromorphic hardware-software co-design~\cite{karunaratne2020memory}.

\section{Conclusion}
\label{sec:conclusion}

In this work, we introduced VS-Graph, a vector-symbolic framework for graph classification that combines efficient hyperdimensional operations with multi-hop structural aggregation to achieve a strong accuracy–efficiency trade-off. VS-Graph uses \textit{``Spike Diffusion"} and \textit{``Associative Message Passing"} to construct expressive graph-level representations in a single pass, leading to substantial improvements over prior HDC methods and achieving accuracy that is on par with or superior to standard GNNs on multiple benchmarks.
These gains are obtained without gradient-based optimization, enabling dramatically lower computational cost: across all datasets and GNN baselines, VS-Graph trains over $250\times$ faster on average, while maintaining ultra-low inference latency. We further showed that the method remains robust under aggressive dimensionality reduction, with minimal loss in accuracy. This stability makes VS-Graph especially attractive for memory-constrained scenarios and low-power neuromorphic platforms. Future work includes pursuing neuromorphic co-design, exploring specialized spiking and in-memory accelerators that can fully exploit VS-Graph’s brain-inspired hyperdimensional computations for further efficiency and scalability.

\section*{Acknowledgment}
The research was funded in part by National Science Foundation through award CCF2319619 and Department of Energy through awards DE-SC0025349 and DE-AC05-00OR22725.


\nocite{*}
\bibliographystyle{IEEEtran}
\bibliography{ref}
\end{document}